\PassOptionsToClass{journal,10pt,twocolumn,letterpaper}{IEEEtran}

\documentclass[journal]{IEEEtran}
\usepackage[T1]{fontenc}
\usepackage{amsmath}
\usepackage{amssymb}
\usepackage{cite}
\usepackage{graphicx}
\usepackage{dblfloatfix}
\usepackage{float}
\usepackage{booktabs}
\usepackage{tabularx}
\usepackage{listings}
\usepackage[hyphens]{xurl}
\usepackage[hidelinks]{hyperref}

\hypersetup{unicode=true}
\graphicspath{{content/article/figures/}}
\newcommand{\software}{gdelt-regkg}

\newcommand{\github}{\url{https://github.com/octo-gone/gdelt-regkg/tree/v2.1.2}}
\newcommand{\pypi}{\url{https://pypi.org/project/gdelt-regkg/2.1.2}}
\newcommand{\githubdictlicenses}{\url{https://github.com/octo-gone/gdelt-regkg/blob/v2.1.2/docs/reference/dictionaries.md\#dictionary-licenses}}

\newcommand{\ArticleNote}[1]{%
  \par\smallskip\begingroup
  \footnotesize\raggedright
  \noindent#1\par
  \endgroup
}

\newcommand{\ArticleReferences}{%
  \bibliographystyle{styles/IEEEtran-extended}%
  \bibliography{content/references/references}%
}

\title{Agreement Between Reconstructed and Native GDELT GKG Records: A Field-Level Baseline}
\author{Nikita~A.~Klimin%
\thanks{N.~A.~Klimin is with HSE University, Moscow, Russia (e-mail: naklimin@edu.hse.ru).}%
}

\newcommand{\ArticleIndexTerms}{Agreement evaluation,
dictionary-based sentiment analysis, Global Knowledge Graph, information extraction,
reproducibility.}

\begin{document}
\maketitle

\begin{abstract}
    The Global Knowledge Graph (GKG) of the Global Database of Events,
Language, and Tone (GDELT) characterizes news articles using named
entities, topics, quantitative and temporal features, quotations, tone,
and dictionary-based content measures. Its codebook defines the record
format, but the software that produces the records is not public, so
researchers cannot generate compatible records for articles they collect
themselves. We built an open-source reconstruction, \software, and
measured how closely it agrees with GDELT's archived (native) records
for 193,550 English articles from 8,729 publishers (2020--2025) and
2,500 articles in five other languages. Agreement depends strongly on
the field. English tone, the percentage of positive minus negative
words, has a mean absolute error (MAE) of 1.79 points and a concordance
correlation coefficient of 0.75. Identity agreement (F1) is 0.65 for
persons, 0.42 for organizations, and 0.65 for locations. It is lower for
date components (0.42), complete counts with their objects and locations
(0.07), and quotations (0.07), while GCAM category presence reaches
micro F1 0.84 within the implemented dictionaries. In subsets where
retrieved and native titles and lengths match, tone error falls to 1.47
and organization F1 rises to 0.48, yet most disagreement remains. A
larger gazetteer raises coverage of native places from 89.5\% to 97.1\%
with little change in location F1. Tone computed on English translations
of non-English articles is less accurate (MAE 2.10--2.64) than for
English articles. The results are conditional on article pages that
could still be retrieved, and the correspondence subsets do not isolate
the effect of input differences. Reconstructed records should therefore
be assessed field by field and not treated as completely interchangeable
with native ones.

\end{abstract}

\section{Introduction}
\label{sec:introduction}
\IEEEPARstart{G}DELT (Global Database of Events, Language, and Tone) is
a machine-coded collection of geolocated news-derived data with global
coverage \cite{leetaru2013gdelt}. One of its main databases is the
Global Knowledge Graph (GKG), which represents news through themes,
sentiment, entities, locations, counts and amounts, and dictionary-based
content measures (Global Content Analysis Measures, GCAM). The GKG
codebook \cite{leetaru2015gkgcodebook} specifies the record format and
field definitions, but formatting alone does not reproduce the
extraction decisions behind original GKG values. Compatible records
would let researchers process independently collected articles through
existing GKG workflows.

This study asks how much of GKG can be reproduced from public
documentation and resources. We built \software, an open-source Python
library, to reconstruct GKG fields from article text. We ran it on
retrieved text of articles that GDELT had processed, and then compared
its output with GDELT's archived output field by field. The codebook and
many of the lexical resources we used are public. Two components,
however, were derived from GDELT;s own output: the gazetteer was built
from original records, and the tone word lists were tuned against
original scores.

There are two sources of uncertainty that affect the comparison: our
implementation differs from GDELT's unreleased pipeline, and the
retrieved text may differ from the text GDELT originally processed. We
therefore ask four questions. First, how does agreement vary across GKG
fields and comparison units? Second, how does agreement change when the
retrieved text is more likely to match the text GDELT processed? Third,
how does agreement change over time, given that older pages are more
likely to have been revised or lost? Fourth, how does agreement vary
with gazetteer size and between original-language processing and
processing of English translations?

\section{About the Software}
\label{sec:about}
\software\ is an open-source Python library and command-line tool for
reconstructing GDELT Global Knowledge Graph fields from news articles.
The library provides a modular pipeline that combines natural language
processing, named-entity recognition and geographic resolution with
compiled dictionaries and extraction rules. Its outputs include themes,
people, organizations, locations, counts, amounts, dates, quotations,
tone, and Global Content Analysis Measures (GCAM). The package also
provides translation, resource compilation and a benchmark for measuring
field-level agreement with original GKG outputs.
\software\ \cite{klimin2026gdeltregkg} is distributed through
PyPI\footnote{\pypi}, and its source code is available at
GitHub\footnote{\github}.

Table~\ref{tab:record-example} illustrates selected outputs for an
article on EU--UK trade negotiations from 41nbc.com (native record
\texttt{20201224001500-1294}). The article was chosen to illustrate the
record format and the comparison; it does not indicate the quality of
the reconstruction in general, which is quantified in
Section~\ref{sec:results}. In this pair, the themes match exactly and
the reconstruction returns one additional ``U.K.'' location entry.

\par\medskip
\noindent\begin{minipage}{\columnwidth}
  \makeatletter\def\@captype{table}\makeatother
  \caption{Example of Native and Reconstructed Fields}
  \label{tab:record-example}
  \centering\footnotesize
  \setlength{\tabcolsep}{3pt}
  \begin{tabularx}{\columnwidth}{@{}l>{\raggedright\arraybackslash}X>{\raggedright\arraybackslash}X@{}}
    \toprule
    Field & Native GDELT & Reconstructed \\
    \midrule
    Persons & boris johnson & boris johnson \\
    Organizations & european union & european union \\
    Location & U.K.; Brussels & U.K.; Brussels; U.K. \\
    Themes & \texttt{LEADER}; \texttt{TAX\_FNCACT}; \ldots & \texttt{LEADER}; \texttt{TAX\_FNCACT}; \ldots \\
    Tone & $-1.471$ & $-1.538$ \\
    Token count & 124 & 130 \\
    \bottomrule
  \end{tabularx}
\end{minipage}\par\medskip

\section{Related Work}
\label{sec:related-work}
An early critique of GDELT highlighted the lack of access to original
source texts, which limited independent verification of coding
decisions, and raised concerns about the conceptual definition of tone
\cite{weller2014raining}. Retrieving articles through their URLs allows
independent examination of GDELT outputs, but the retrieved text may
differ from what GDELT originally processed.

Later assessments raised further concerns about data quality. The UK
Office for National Statistics noted that it remains unclear whether
specific quality-assurance mechanisms are incorporated into GDELT's data
processing pipeline \cite{ons2020gdelt}. Wang et al.
\cite{wang2016growing} identified reliability and validity concerns in a
comparative evaluation of automated event databases, including GDELT and
ICEWS. Their analysis revealed a considerable reduction in the number of
usable events following URL retrieval, content verification,
deduplication, and event classification. Specifically, only 21\% of
GDELT records with valid URLs were found to correspond to the target
events. A more recent statistical assessment of the GDELT event database
reported approximately 55\% accuracy across key fields and a redundancy
rate of around 20\% \cite{hong2025gdelt}. Only one of the studies above
concerns GKG directly: an examination of 100 randomly selected news
articles found that GDELT's theme extraction algorithm tended to
identify themes even when none were present
\cite[p.~3]{tilly2021macroeconomic}.

The event-database assessments do not directly establish the quality of
GKG fields, and no conclusions about GKG can be inferred from them. It
is also worth noting that the studies above primarily evaluated archived
records as given or checked GDELT against external sources, rather than
reproducing its outputs. Certain GKG fields may be more amenable to
partial reproduction than the event database, because their underlying
methods are well described in the GKG
codebook~\cite{leetaru2015gkgcodebook} and in GDELT's technical
posts\footnote{https://blog.gdeltproject.org}. Nevertheless, reproducing
GDELT's outputs remains ambitious even with improved processing methods,
as undocumented algorithms may prevent exact replication.

Kohlhepp~\cite{kohlhepp2025kngdelt} proposes K-N-GDELT as a replacement
for GDELT, rather than a reconstruction of its native outputs. The
system uses dictionary-based event coding and large language models and
does not aim for backward compatibility with GDELT records. The report
presents geopolitical applications but gives limited implementation
details and no quantitative evaluation of extraction accuracy or of
agreement with native GKG fields, so its outputs cannot be compared with
ours. Its claims of improved extraction are therefore not established by
the reported evidence.

Reconstructing native outputs also depends on the accessibility and
mutability of online articles. Link rot can make previously available
articles inaccessible, potentially eliminating the evidence needed to
reproduce earlier findings, with more effect on older pages
\cite{chapekis2024disappears}. Even when URLs remain accessible, their
content may change over time due to publisher revisions or website
practices that reuse existing URLs for updated content. Research on news
revisions identifies factual, stylistic, and narrative changes,
demonstrating that an article URL does not necessarily identify an
immutable version of its content~\cite{spangher2024newsedits2}.
Furthermore, differences in article-body extraction may produce
inconsistent representations of the same source, introducing additional
variability before downstream analysis.

Web-processing tools also introduce such variability. For example,
news-please~\cite{hamborg2017newsplease} combines extraction tools and
heuristics to recover article text and metadata;
Trafilatura~\cite{barbaresi2021trafilatura} targets the main text while
removing irrelevant elements; and simpler tools, such as Beautiful
Soup~\cite{richardson2025beautifulsoup}, provide HTML parsing, element
selection and visible text extraction. These approaches can retain
different paragraphs, captions, and other elements from the same page,
potentially affecting later processing.

To summarize, reproducing and evaluating GKG outputs remains challenging
without access to the original processing pipeline and the exact source
texts used by GDELT. The available information provides a basis for
partial reconstruction but does not fully specify the details needed for
faithful replication. To the best of our knowledge, \software\ is the
first public package specifically designed to reconstruct GKG-compatible
records from article text. It enables researchers to process
independently collected articles using the GKG-compatible format. This
study presents the package and evaluates its agreement with native GKG
records field by field.

\section{Materials}
\label{sec:materials}
In this work, \emph{native} denotes values in archived GKG records.
\emph{Reconstructed} denotes values computed from the retrieved bodies.
For numeric comparisons, $y_i$ is native and $\widehat y_i$ is
reconstructed. The article--native-record pair is the evaluation unit.

\subsection{English Corpus}
\label{sec:english-corpus}
English candidates came from six scheduled GKG archive windows per month
between 1 January 2020 and 31 December 2025; unavailable windows were
not replaced. GKG is published as archive files every 15 minutes; each
scheduled archive is one window. Records were parsed from each retrieved
archive file. A record was eligible if it came from the web source
collection, had an HTTP(S) URL, and was not marked as translated. For a
month with $D$ days, sampling selected days $\lceil D/4\rceil$ and
$\lceil3D/4\rceil$, at 00:15, 12:00, and 23:45 UTC. Across the 72-month
period, 421 of the 432 scheduled archives were available and 11 were
unavailable. Repeated URLs were consolidated; conflicts in
extraction-relevant native fields were audited before unambiguous
pairing.

Article pages were retrieved on 8--9 October 2026 (UTC), approximately 9
months to 7 years after the publication date of the archives. A URL was
counted as failed if the download or processing raised an error (network
or HTTP errors, non-HTML responses, responses larger than 5 MiB), if no
text could be extracted, if the extracted body had fewer than 50 tone
tokens, or if the page was detected as an access-denied or
browser-challenge notice. Bodies not identified as English by the
language detector were then excluded.

Titles were stored separately and were not prepended to the body. Exact
duplicates of normalized token sequences and near duplicates at a
Hamming distance $\leq 3$ bits of a 64-bit SimHash were removed
\cite{manku2007detecting}.

Annual acquisition and retention are summarized in
Table~\ref{tab:acquisition-retention}.

\begin{table}[!htbp]
  \caption{Acquisition and Retention}
  \label{tab:acquisition-retention}
  \centering
  \begin{tabular*}{\columnwidth}{@{\extracolsep{\fill}}lrrr@{}}
    \toprule
    Year & Attempted URLs & Retained & Retention \(\%\) \\
    \midrule
    2020 & 105,772 & 29,068 & 27.48 \\
    2021 & 92,123 & 25,144 & 27.29 \\
    2022 & 83,638 & 25,572 & 30.57 \\
    2023 & 107,833 & 36,957 & 34.27 \\
    2024 & 99,995 & 39,351 & 39.35 \\
    2025 & 87,860 & 37,458 & 42.63 \\
    \midrule
    \textbf{Total} & \textbf{577,221} & \textbf{193,550} & \textbf{33.53} \\
    \bottomrule
  \end{tabular*}
\end{table}

\subsection{Multilingual Corpus}
\label{sec:multilingual-corpus}
The multilingual sample used translated GKG archives corresponding to
those used for the English sample; no other archive windows were
requested. The target was 500 original-language articles each in French,
Spanish, German, Russian, and Chinese, stratified across 2020--2025. The
original language came from the native translation-information field and
was checked against the retrieved body using language detector. This
collection required at least 250 alphabetic Unicode characters, with
exact-body deduplication and a cap of 50 articles per hostname within
each language.

Of the 421 corresponding translated archives, 420 contained required
entries. Annual selection targets were 84 articles per language in each
of 2020 and 2021 and 83 in each remaining year. We generated English
translations of the full texts of all French, Spanish, German, and
Russian articles and of 499 Chinese articles. One Chinese translation
reached the generation limit and was excluded from translation-dependent
comparisons.

The English and multilingual samples have separate inclusion rules.
Their processing is described in Section~\ref{sec:methods-gcam} and
their results in Section~\ref{sec:results-multilingual}.

\section{Methods}
\label{sec:methods}

\subsection{Software}
\label{sec:methods-software}
English acquisition used Trafilatura
2.3.0~\cite{barbaresi2021trafilatura} to extract the body and title,
with comments and tables excluded. Lingua 2.2.0 \cite{stahl2026lingua}
identified the language. We generated English translations with the
\texttt{facebook/m2m100\_418M} checkpoint of
M2M100~\cite{fan2021m2m100}, using Transformers
5.19.0~\cite{wolf2020transformers}, CUDA, and half-precision weights.

GKG fields were reconstructed using \software\ 2.1.2
\cite{klimin2026gdeltregkg}. Entity spans were computed once per English
body and reused across geographic conditions. The frozen resource
profiles were \texttt{tone.v3}, \texttt{themes.v3}, \texttt{counts.v3},
\texttt{names.v1}, and custom GCAM \texttt{v5} with supplied
LIWC~\cite{pennebaker2007liwc} and MPQA
Subjectivity~\cite{wilson2005polarity} inputs. These labels identify
versions of the extraction resources, which are versioned separately
from the package release. Named-entity recognition used spaCy
3.8.16~\cite{honnibal2020spacy} and en\_core\_web\_lg
3.8.0~\cite{explosion2024encoreweblg}. LIWC is licensed and is not
redistributed. Each dictionary is computed independently, so the results
for the other dictionaries do not depend on it. Dictionary licenses are
listed in the \software\
 documentation\footnote{\githubdictlicenses}.

\subsection{Tone}
\label{sec:methods-tone}
Tone used a General Inquirer-based vocabulary~\cite{stone1966inquirer}.
Positive, negative, activity, and self/group word lists were refined
against GKG scores in a separate development corpus built from native
GKG files that do not overlap the English corpus, from
September--October 2026. It contained 1,059 training articles and 1,322
validation articles. Candidate word additions and removals were fitted
on the training articles and selected on the validation articles to
reduce disagreement with native tone and its components.The lists were
fixed before benchmark evaluation.

The tone tokenizer produces normalized word and number tokens while
retaining internal apostrophes and hyphens. Tone was approximated as
\begin{equation}
    \label{eq:reconstructed-tone}
    \widehat T_i=100\frac{m_i^+-m_i^-}{N_i^{\mathrm{rec}}}
\end{equation}
where $N_i^{\mathrm{rec}}$ is the reconstructed token count and $m_i^+$
and $m_i^-$ count positive and negative lexical matches. Repeated tokens
and stopwords remain in $N_i^{\mathrm{rec}}$. Positive Score and
Negative Score are the percentages of tokens matching the corresponding
word lists. Polarity uses the union of the positive and negative lists.
Activity Reference Density and Self/Group Reference Density use their
respective word lists and the same denominator. The GKG Word Count
component is represented by $N_i^{\mathrm{rec}}$. These are
reconstructed approximations of the corresponding GKG tone components.

\subsection{Geographic Resources}
\label{sec:geographic-resources}
The primary configuration used an \emph{expanded} gazetteer with 119,733
entries. The comparison used the \emph{bundled} inventory of 17,177
entries, built with the default \software\ configuration. Both were
built from native GKG location fields dated 2015 to before 2020 without
use of external gazetteers, and remained fixed during evaluation. The
same English bodies, cached entity spans, and extraction rules were used
for both inventories. A larger inventory adds available identities,
spelling aliases, and historical observation frequencies, which should
help resolve ambiguous place names.

For native location identities $L_i$ in article $i$ and gazetteer
inventory $G$, assignment coverage is
\begin{equation}
    \label{eq:gazetteer-coverage}
    C(G)=\frac{\sum_i |L_i\cap G|}{\sum_i |L_i|}.
\end{equation}
Each identity contributes once per article. Coverage measures whether
the resource contains native identities; location F1 measures whether
the reconstruction recovers them. The resource contrast is the paired
expanded-minus-bundled difference. We also evaluate complete count
tuples, which require geographic attachments to agree.

\subsection{GCAM}
\label{sec:methods-gcam}
This work defines a category as \emph{implemented} if it has a rule or
lexical resource in the reconstruction profile, and as \emph{observed}
if it occurs in at least one native or reconstructed output within the
evaluated subset. Thus, some implemented categories may never appear in
the native outputs, while some observed categories may appear only in
native outputs and have no corresponding implementation.

The frozen GCAM profile covers 2,492 categories across the evaluated
languages, identified by 31 dictionary IDs. Lexical-match dictionaries
count matched token or phrase occurrences within each category. Weighted
dictionaries average the imported word ratings over matched token
occurrences. The profile includes 268 authored GKG theme-category
approximations and one French imagery approximation based on
rating-interval midpoints. Table~\ref{tab:gcam-resources} and
Appendix~\ref{app:resources} document all dictionaries and procedures
used.

We followed GDELT's distinction between original-language GCAM and
analysis of English translations~\cite{gdelt2015translingual} and kept
processing separate. English dictionaries were applied to the English
corpus and to translations. Original-language dictionaries were applied
only to the corresponding original-language bodies.

Category \emph{presence} means a lexical count greater than zero.
Presence is evaluated separately from weighted-score availability and
agreement on score values. A \emph{companion count} is the number of
lexical matches associated with a weighted score. If a native weighted
score is reported without a companion count, the corresponding
article--category pair is excluded from both the native and
reconstructed outputs used in the comparison.

A reconstructed-only assignment occurs when an article--category pair
has a positive reconstructed count but no positive native count in the
retained comparison. Native count support is the number of
article--category pairs with a positive native count.

\subsection{Input Correspondence}
\label{sec:input-correspondence}
The retrieved bodies may differ from the text processed by GDELT. We
examined agreement in subsets selected by title equality, body length,
or quotation anchors. Each subset used the same reconstructed outputs as
the full English corpus.

The native title came from the \texttt{PAGE\_TITLE} element of the GKG
extra-XML field, with embedded tags removed. The reconstructed title
came from the stored Trafilatura extraction metadata. Both were
HTML-unescaped, Unicode-NFKC normalized, case-folded, and split into
alphanumeric tokens. The \emph{Title-available} subset required nonempty
token sequences on both sides. The \emph{Title-matched} subset required
both titles to contain at least four tokens and at least three distinct
tokens outside the fixed stopword list. It also required equality of the
complete normalized token sequences; stopwords were retained in that
equality check. These criteria defined diagnostic subsets only.

Length correspondence used native and reconstructed tone token counts:
\begin{equation}
    \label{eq:length-ratio}
    r_i=\frac{N_i^{\mathrm{rec}}}{N_i^{\mathrm{native}}}.
\end{equation}
The native denominator had to be positive. The
\emph{Title-and-length ($\mathit{\pm10\%}$) subset} required Title-matched
eligibility and $0.90\leq r_i\leq1.10$. Sensitivity analyses used length
bands of 5\% and 20\%.

A quotation-anchor subset was selected independently of title and
length. An article required at least two distinct eligible native
quotations; an eligible quotation contained at least 40 characters and
eight whitespace-separated tokens. At least 80\% of its eligible
quotations had to occur in the normalized retrieved body.

We assessed sensitivity to additional near-duplicate removal by applying
a five-bit SimHash Hamming-distance cutoff to the retained English
corpus. Agreement was recalculated using the existing predictions. This
comparison assesses sensitivity to corpus composition.

\begin{table*}[!b]
  \caption{Field Reconstruction and Comparison Units}
  \label{tab:field-comparisons}
  \centering\footnotesize
  \setlength{\tabcolsep}{4pt}
  \begin{tabularx}{\textwidth}{@{}>{\raggedright\arraybackslash}p{0.17\textwidth}>{\raggedright\arraybackslash}p{0.25\textwidth}>{\raggedright\arraybackslash}X@{}}
    \toprule
    Field & Reconstruction & Matched unit within an article \\
    \midrule
    Persons, organizations, all names & Named-entity spans and name rules & Sets of normalized name identities \\
    Themes & Phrase, wildcard, and context rules & Sets of exact theme codes \\
    Locations & Place recognition and gazetteer resolution & Type, country, first-level administrative code (ADM1), and feature ID \\
    Counts & Quantity--object--location templates & Multisets: \emph{label + quantity}; \emph{object} additionally requires the attached object text; \emph{full tuple} also requires location type, name, country, ADM1, latitude, longitude, and feature ID \\
    Dates & Calendar-validated patterns & Resolution and available month, day, and year components \\
    Amounts & Numeric/scale normalization and object attachment & Sets of numeric \emph{values}; \emph{value + object} additionally requires the same normalized attached object text \\
    Quotations & Delimiters and reporting-verb rules & Sets of normalized quotation texts \\
    Tone & General Inquirer-based vocabulary & Paired scalar scores, Eq.~\eqref{eq:reconstructed-tone} \\
    GCAM & Language-specific dictionaries & Lexical-match category presence; implemented lexical counts; paired weighted scores \\
    \bottomrule
  \end{tabularx}
\end{table*}

\subsection{Agreement Measures}
\label{sec:agreement-measures}
For article $i$, within-article set or multiset matching gives true
positives $TP_i$, false positives $FP_i$, and false negatives $FN_i$.
The principal categorical endpoint pools these counts:
\begin{equation}
    \label{eq:micro-f1}
    F_{1,\mathrm{micro}}=
    \frac{2\sum_i TP_i}{2\sum_i TP_i+\sum_i FP_i+\sum_i FN_i}.
\end{equation}
An item cannot match across articles. Precision divides pooled true
positives by the number of reconstructed items; recall divides them by
the number of native items.

For a set $U_+$ of units with nonzero F1 denominators, macro F1 is
\begin{equation}
    \label{eq:macro-f1}
    F_{1,\mathrm{macro}}=\frac{1}{|U_+|}\sum_{u\in U_+}
    \frac{2TP_u}{2TP_u+FP_u+FN_u}.
\end{equation}
Both-empty units are excluded from $U_+$; units with content on only one
side contribute zero.

The unit $u$ is an article, a category, or a publisher for
\mbox{article-}, \mbox{category-}, or publisher-macro F1. Counts are
pooled within each category or publisher before averaging.

Category-macro F1 covers theme labels and GCAM lexical-match categories,
separately for all observed categories and the applicable implemented
inventory. Article-macro F1 has cluster intervals for the principal
endpoints.

For $n$ valid numeric pairs, we report error in the original score
units:
\begin{equation}
    \label{eq:mae}
    \mathrm{MAE}=\frac{1}{n}\sum_{i=1}^{n}|\widehat y_i-y_i|.
\end{equation}
\begin{equation}
    \label{eq:bias}
    \mathrm{Bias}=\frac{1}{n}\sum_{i=1}^{n}(\widehat y_i-y_i).
\end{equation}
\begin{equation}
    \label{eq:rmse}
    \mathrm{RMSE}=\sqrt{\frac{1}{n}\sum_{i=1}^{n}(\widehat y_i-y_i)^2}.
\end{equation}
Lin's concordance correlation coefficient~\cite{lin1989concordance} is
\begin{equation}
    \label{eq:ccc}
    \mathrm{CCC}=\frac{2\operatorname{Cov}(Y,\widehat Y)}
    {\operatorname{Var}(Y)+\operatorname{Var}(\widehat Y)
        +(\bar y-\overline{\widehat y})^2}.
\end{equation}
Here $\bar y=n^{-1}\sum_i y_i$ and
$\overline{\widehat y}=n^{-1}\sum_i\widehat y_i$ are the paired-sample
means, and $\operatorname{Var}(Y)=n^{-1}\sum_i(y_i-\bar y)^2$, with an
analogous definition for $\widehat Y$, and
$\operatorname{Cov}(Y,\widehat Y)=n^{-1}\sum_i(y_i-\bar y)(\widehat y_i-\overline{\widehat y})$.
The same divisor $n$ is used throughout. CCC is undefined for
insufficient or numerically degenerate paired samples.

As a separate descriptive ratio, tone MAE was divided by the sample
standard deviation of native tone $s_{\text{native}}$. The ratio does
not replace MAE and expresses error relative to variability in native
tone.

Tone sign agreement uses a neutral band with $\tau=0.5$:
\begin{equation}
    \label{eq:tone-sign}
    s_\tau(z)=\begin{cases}
        -1, & z<-\tau,             \\
        0,  & -\tau\leq z\leq\tau, \\
        1,  & z>\tau.
    \end{cases}
\end{equation}
\begin{equation}
    \label{eq:sign-agreement}
    \mathrm{SignAgreement}=\frac{1}{n}\sum_{i=1}^{n}
    \mathbf{1}\!\left[s_\tau(\widehat y_i)=s_\tau(y_i)\right] \times 100\%.
\end{equation}
The Bland--Altman plot displays $\widehat y_i-y_i$ against
$(y_i+\widehat y_i)/2$. The descriptive 95\% limits of agreement are
calculated as the bias $\pm1.96$ sample standard deviations (SD) of the
differences~\cite{bland1986agreement}.

Weighted absolute percentage error (WAPE) is used for implemented GCAM
families with native lexical-count support:

\begin{equation}
    \label{eq:gcam-wape}
    \mathrm{WAPE}_d=
    \frac{\sum_{i=1}^{n}\sum_{k\in K_d}|\widehat y_{ik}-y_{ik}|}
    {\sum_{i=1}^{n}\sum_{k\in K_d}y_{ik}} \times 100\%.
\end{equation}
Here $y_{ik}$ and $\widehat y_{ik}$ are the native and reconstructed
occurrence counts for article $i$ and implemented category $k$, and
$K_d$ contains the implemented comparable lexical-match categories in
family $d$. WAPE is undefined when the native-count denominator is not
positive.

For weighted scores, availability F1 was

\begin{equation}
    \label{eq:availability-f1}
    F_{1,\mathrm{availability}}=\frac{2c_{\mathrm{both}}}{c_{\mathrm{native}}+c_{\mathrm{rec}}},
\end{equation}
where the counts denote article--category pairs with both values
($c_{\mathrm{both}}$), a native value ($c_{\mathrm{native}}$), or a
reconstructed value ($c_{\mathrm{rec}}$). An available value of zero was
retained; a missing value was not. Paired-value errors used only pairs
with both values.

Perfect agreement gives F1 and CCC of 1, sign agreement of 100\%, and
MAE, RMSE, and WAPE of 0. Zero bias alone does not establish agreement.

\subsection{Alignment}
\label{sec:alignment}
Outputs were aligned by document identifier, timestamp, and source
collection before scoring. Name normalization used Unicode NFKC, case
folding, apostrophe normalization, and whitespace collapse while
preserving punctuation. Theme codes were compared exactly. The
standalone Themes field used 519 implemented labels and was evaluated
separately from the 367 GKG Themes categories represented within GCAM.
Count comparisons retained tuple multiplicity. Malformed native fields
were excluded only from the affected comparison; malformed categorical
predictions were treated as empty, retaining their native items as false
negatives. Numeric errors required valid paired values. Valid empty
fields and unavailable values were separated throughout. For categorical
comparisons, valid articles have a well-formed native value for the
field, while \emph{active} articles contain at least one item in either
output.

Table~\ref{tab:field-comparisons} defines the matching units used in the
field-level results, including the progressively stricter count
comparisons.

\begin{table*}[!b]
  \caption{English Field Agreement}
  \label{tab:english-field-agreement}
  \centering
  \footnotesize
  \setlength{\tabcolsep}{3pt}
  \begin{tabular*}{\textwidth}{@{\extracolsep{\fill}}lrrrrr@{}}
    \toprule
    Field / comparison & Active articles & Micro F1 & Article-macro F1 & Category-macro F1 & Publisher-macro F1 \\
    \midrule
    Persons / normalized identities & 172,393 & 0.646 [0.637, 0.656] & 0.592 [0.582, 0.603] & --- & 0.589 \\
    Organizations / normalized identities & 158,843 & 0.417 [0.411, 0.422] & 0.325 [0.319, 0.331] & --- & 0.357 \\
    Locations / identities & 167,617 & 0.654 [0.646, 0.662] & 0.543 [0.532, 0.554] & --- & 0.586 \\
    Themes / all labels & 192,585 & 0.617 [0.611, 0.622] & 0.526 [0.519, 0.534] & 0.028 & 0.529 \\
    Counts / label + quantity & 28,143 & 0.285 [0.276, 0.295] & 0.153 [0.147, 0.159] & --- & 0.179 \\
    Counts / object & 28,143 & 0.166 [0.158, 0.176] & 0.088 [0.085, 0.092] & --- & 0.104 \\
    Counts / full tuple & 28,143 & 0.070 [0.066, 0.076] & 0.039 [0.037, 0.041] & --- & 0.046 \\
    Dates / components & 140,402 & 0.424 [0.409, 0.439] & 0.308 [0.293, 0.322] & --- & 0.366 \\
    Amounts / values & 181,720 & 0.406 [0.397, 0.415] & 0.377 [0.370, 0.383] & --- & 0.396 \\
    Amounts / value + object & 181,720 & 0.198 [0.193, 0.203] & 0.180 [0.177, 0.184] & --- & 0.195 \\
    All names / identities & 193,504 & 0.231 [0.226, 0.236] & 0.225 [0.220, 0.230] & --- & 0.220 \\
    Quotations / content & 148,422 & 0.068 [0.063, 0.074] & 0.053 [0.050, 0.055] & --- & 0.063 \\
    Themes / implemented labels & 192,510 & 0.698 & 0.597 & 0.675 & --- \\
    GCAM / presence & 193,550 & 0.824 [0.819, 0.827] & 0.809 [0.805, 0.814] & 0.651 & 0.811 \\
    GCAM / implemented presence & 193,550 & 0.840 [0.836, 0.844] & 0.826 [0.821, 0.830] & 0.674 & 0.828 \\
    \bottomrule
  \end{tabular*}
  \ArticleNote{Active articles have at least one item in either output. Micro F1 pools within-article matches; article-macro F1 averages over articles with at least one item on either side. Category-macro F1 applies to theme labels and GCAM categories observed in either output. Publisher-macro F1 averages defined publisher-specific scores. Brackets give 95\% publisher-cluster intervals where calculated. A dash denotes an inapplicable or uncalculated average.}
\end{table*}

\subsection{Temporal Agreement}
\label{sec:methods-temporal}
English pairs were grouped into 72 months by native GKG timestamp, using
the expanded gazetteer throughout. Each month is sampled on two days, so
each calendar month is represented by articles represented by 12 sampled
days across the six years. We computed monthly tone MAE, micro F1, and
dictionary-specific count WAPE for the full corpus, the Title-matched
subset, and the Title-and-length subsets with 5\%, 10\%, and 20\% bands.
Comparisons spanning several months pooled matching counts and WAPE
numerators and denominators; tone MAE was weighted by the number of
valid pairs. Period boundaries were chosen after inspecting the
full-corpus monthly series and are therefore exploratory; they were
applied unchanged to all subsets and dictionaries. To examine recurring
calendar patterns, date-component agreement was compared by year and
date resolution, then pooled by calendar month across years. Temporal
comparisons are descriptive point estimates.

\subsection{Aggregation and Uncertainty}
\label{sec:uncertainty}
Six-year estimates were recalculated from pooled article contributions,
not averaged across monthly scores. Publishers were grouped by native
source-common-name, with URL hostname as fallback. Principal intervals
used 2,000 publisher-cluster bootstrap replicates with percentile limits
at 2.5\% and 97.5\%~\cite{field2007bootstrap}. Resampling a publisher
retained all its article contributions together. English conditions used
a common publisher frame and identical draws; paired contrast intervals
used replicate-wise differences. Multilingual analyses used separate
language-and-stream frames. Numeric publisher counts included only
sources with valid pairs. Multilingual endpoints with fewer than 100
paired articles or 20 contributing publishers were flagged as
descriptive.

\section{Results}
\label{sec:results}

\subsection{Corpus and Attrition}
\label{sec:results-corpus}
The English collection retained 193,550 articles from 8,729 publishers.
Of 577,221 attempted unique URLs, 311,455 failed the retrieval and
extraction checks described in Section~\ref{sec:materials}, and 34,687
were not identified as English. Of the remaining 231,079 articles,
33,218 were exact-body duplicates, 4,308 were near duplicates, and 3 had
conflicting native records; these were excluded.

The annual retention fraction rose from 27.5\% in 2020 to 42.6\% in 2025
(Table~\ref{tab:acquisition-retention}). This retention measure combines
retrieval, extraction, language screening, and duplicate and
native-record exclusions.

\subsection{Categorical Field Agreement}
\label{sec:results-english}
Person and location identities had higher agreement than organizations,
whereas complete counts and quotation texts had the lowest F1 among the
principal comparisons (Table~\ref{tab:english-field-agreement}). Person,
organization, and location micro F1 were 0.646 [0.637, 0.656], 0.417
[0.411, 0.422], and 0.654 [0.646, 0.662], respectively. The table
reports the number of active articles for each comparison, meaning
articles with at least one item in either output, rather than the total
corpus size. All valid articles contribute to pooled matching;
both-empty pairs add no items. Location and date comparisons had 193,502
and 193,541 valid articles, respectively; the other rows used all
193,550 articles.

Person precision and recall were 0.571 and 0.745; organization precision
and recall were 0.492 and 0.362. Article-macro F1 was 0.592 [0.582,
0.603] for persons and 0.325 [0.319, 0.331] for organizations, below
their pooled micro scores. Publisher-macro F1 was 0.589 and 0.357,
respectively.

Counts lost agreement as the required tuple became more specific: F1 was
0.285 [0.276, 0.295] for label--quantity pairs, 0.166 [0.158, 0.176]
after requiring the attached object, and 0.070 [0.066, 0.076] after also
requiring all geographic fields. Amount-value F1 was 0.406 [0.397,
0.415], falling to 0.198 [0.193, 0.203] when the object also had to
match. These comparisons use the distinct units in
Table~\ref{tab:field-comparisons}; matching a number alone does not
imply agreement on what it describes.

Date-component F1 was 0.424 [0.409, 0.439], with precision 0.290 and
recall 0.792. The component-specific results varied from F1 0.652 for
day--month dates to 0.125 for day--month--year dates. Quotation-text F1
was 0.068 [0.063, 0.074], and the broader all-names field had F1 0.231
[0.226, 0.236].

Theme micro F1 was 0.617, but category-macro F1 was 0.028 across 12,597
observed labels. Restricting both outputs to implemented labels gave
micro F1 0.698 and category-macro F1 0.675 across 519 labels.

\subsection{Tone Agreement}
\label{sec:results-tone}
\begin{figure}[!htbp]
  \centering
  \includegraphics[width=\columnwidth]{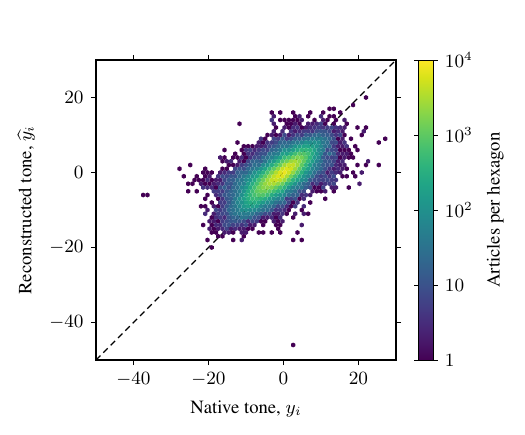}
  \caption{Native and reconstructed tone for all 193,550 English articles.
    Color shows the number of articles per hexagon. The dashed line is identity.}
  \label{fig:tone-agreement}
\end{figure}

Tone MAE was 1.79 [1.76, 1.82] and CCC was 0.749 [0.736, 0.759] across
all 193,550 English pairs. Native tone had sample standard deviation
$s_{\text{native}}$ of 3.891, giving $\mathrm{MAE}/s_{\text{native}}$
0.459. Signed bias was 0.22 points, RMSE was 2.50, and three-class sign
agreement was 70.5\%. Thus the small average signed difference coexisted
with appreciable article-level absolute errors.
Fig.~\ref{fig:tone-agreement} compares the paired tone values with the
identity line.

\begin{figure}[!htbp]
  \centering
  \includegraphics[width=\columnwidth]{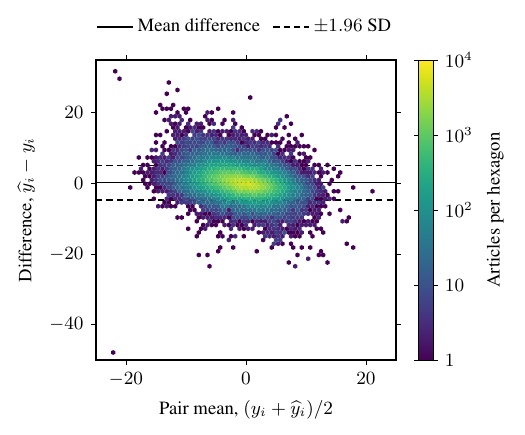}
  \caption{Tone differences against the pair mean for all 193,550 English articles.
    Color shows articles per hexagon. The solid line is mean
    bias (0.22); dashed lines are bias $\pm1.96$ sample SD of the
    differences ($-4.67$, 5.10).}
  \label{fig:tone-differences}
\end{figure}

The Bland--Altman plot shows the reconstructed-minus-native differences
against the pair mean (Fig.~\ref{fig:tone-differences}). The
descriptive limits were $-4.67$ and 5.10 points. Both plots show all
English pairs, including extreme values.

For the remaining tone components, positive and negative density had MAE
1.07 and 1.34 points and CCC 0.664 and 0.702. Activity density had MAE
3.19 points and bias $-1.58$ points. Reconstructed token counts exceeded
native counts by 81.3 tokens on average, with MAE 147.1 tokens and CCC
0.638. Polarity and self/group density had MAE 1.75 and 0.35 points and
CCC 0.581 and 0.705, respectively. These components retain their own
scales; their MAE values are not directly comparable across units.

\subsection{Input Correspondence}
\label{sec:results-input-correspondence}
Title correspondence alone was associated with little change in tone
error, while the Title-and-length ($\pm10\%$) subset had lower tone
error and higher entity F1
(Table~\ref{tab:agreement-input-correspondence}). The Title-available
subset contained 193,100 articles, and the Title-matched subset
contained 117,217. Tone MAE was 1.79 in the Title-available subset and
1.77 in the Title-matched subset, compared with 1.79 in the full corpus.

\begin{table*}[!p]
\caption{Agreement by Input Correspondence and Deduplication}
\label{tab:agreement-input-correspondence}
\centering\footnotesize
\setlength{\tabcolsep}{4pt}
\begin{tabular*}{\textwidth}{@{\extracolsep{\fill}}lrrrrrr@{}}
\toprule
Subset & Articles & Tone MAE & Tone CCC & Person F1 & Org. F1 & Location F1 \\
\midrule
Full corpus & 193,550 & 1.79 [1.76, 1.82] & 0.749 [0.736, 0.759] & 0.646 & 0.417 & 0.654 \\
Title-available & 193,100 & 1.79 [1.76, 1.82] & 0.749 [0.737, 0.759] & 0.647 & 0.417 & 0.654 \\
Title-matched & 117,217 & 1.77 [1.73, 1.81] & 0.759 [0.742, 0.772] & 0.664 & 0.435 & 0.671 \\
Title-and-length ($\pm5\%$) subset & 23,121 & 1.48 [1.46, 1.51] & 0.827 [0.822, 0.833] & 0.736 & 0.473 & 0.720 \\
Title-and-length ($\pm10\%$) subset & 54,485 & 1.47 [1.46, 1.49] & 0.830 [0.826, 0.834] & 0.748 & 0.483 & 0.727 \\
Title-and-length ($\pm20\%$) subset & 81,668 & 1.51 [1.49, 1.52] & 0.830 [0.827, 0.833] & 0.742 & 0.475 & 0.723 \\
Quotation anchors & 1,067 & 1.41 [1.35, 1.48] & 0.810 [0.786, 0.832] & 0.735 & 0.479 & 0.746 \\
Stricter deduplication ($\leq5$ bits) & 192,135 & 1.78 [1.75, 1.82] & 0.750 [0.738, 0.760] & 0.649 & 0.418 & 0.656 \\
\bottomrule
\end{tabular*}
\ArticleNote{F1 is micro-averaged. Brackets give 95\% publisher-cluster intervals for tone.}
\end{table*}

The Title-and-length ($\pm10\%$) subset contained 54,485 articles. Tone
MAE was 1.47 [1.46, 1.49] and CCC was 0.830 [0.826, 0.834]; person,
organization, and location F1 were 0.748, 0.483, and 0.727. The 5\% and
20\% bands retained 23,121 and 81,668 articles, with tone MAE 1.48 and
1.51, respectively. The narrower band did not yield the lowest MAE;
across these bands, organization F1 ranged from 0.473 to 0.483 and
location F1 from 0.720 to 0.727.

The quotation-anchor subset, selected independently of title and length,
contained 1,067 articles, with tone MAE 1.41 [1.35, 1.48], CCC 0.810
[0.786, 0.832], and location F1 0.746.

\subsection{Deduplication Robustness}
\label{sec:results-robustness}
Applying additional duplicate-detection removed further 1,415 articles.
On the remaining 192,135 articles, agreement changed only marginally:
tone MAE was 1.78, and person, organization, and location F1 were 0.649,
0.418, and 0.656 (Table~\ref{tab:agreement-input-correspondence}).

\subsection{Temporal Agreement}
\label{sec:results-temporal}
\begin{figure*}[!p]
  \centering
  \includegraphics[width=\textwidth]{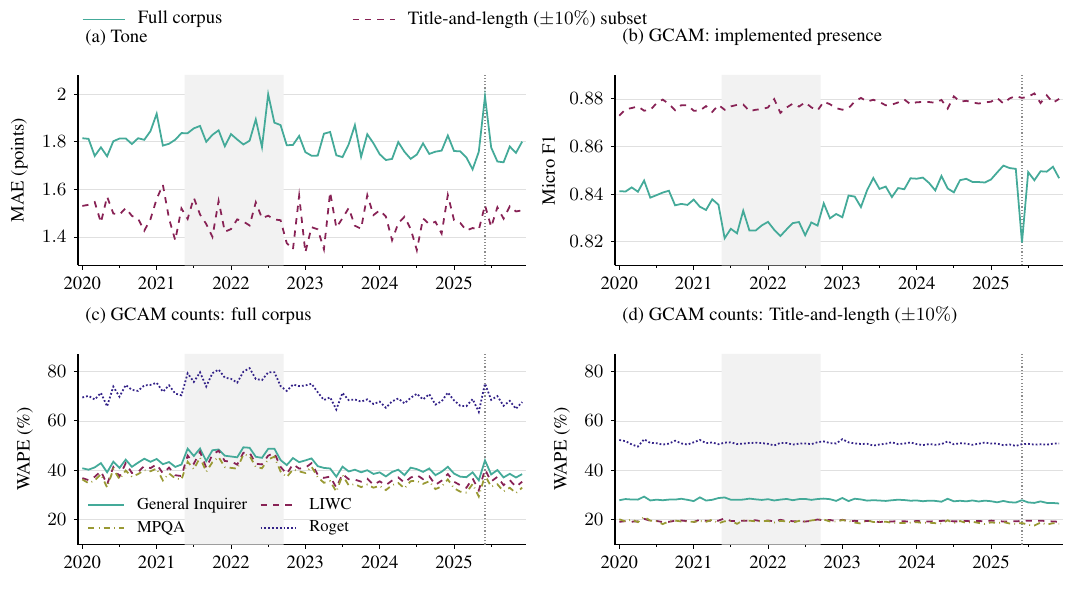}
  \caption{Monthly agreement for the full corpus and the Title-and-length ($\pm10$\%)
  subset: (a) tone MAE; (b) GCAM implemented presence micro F1; (c) count WAPE for four dictionaries,
  full corpus; (d) the same, Title-and-length ($\pm10$\%) subset.
  Points are monthly estimates without uncertainty. Shading marks
  June 2021--September 2022; the dotted line marks June 2025.}
  \label{fig:monthly-agreement}
\end{figure*}

Agreement did not improve consistently with more recent native records
(Fig.~\ref{fig:monthly-agreement}). Full-corpus monthly tone MAE
ranged from 1.68 to 2.00 points. GCAM count disagreement increased
across several dictionaries from approximately June 2021 through
September 2022, then decreased.

General Inquirer count WAPE was 42.2\% in January 2020--May 2021, 46.9\%
in June 2021--September 2022, and 39.7\% in October 2022--May 2025. The
Title-and-length ($\pm10\%$) subset was nearly stable over the same
periods, at 28.2\%, 28.3\%, and 27.7\%. The same pattern held for LIWC,
MPQA, and Roget and under the 5\% and 20\% length bands. The
Title-matched subset retained the 2021--2022 increase.

\begin{figure*}[!p]
  \centering
  \includegraphics[width=\textwidth]{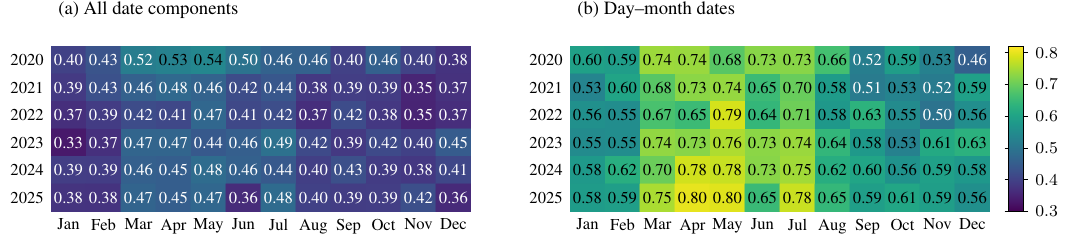}
  \caption{Date-component micro F1 by year (rows) and calendar month (columns):
  (a) all date components;
  (b) day--month dates without a year. Each month is sampled on two days.}
  \label{fig:calendar-agreement}
\end{figure*}

Date agreement showed a recurring calendar pattern in the sampled days
(Fig.~\ref{fig:calendar-agreement}). Pooled date-component F1 was
0.469 for March--May and 0.395 for September--December. The contrast was
larger for day--month dates without a year: 0.742 versus 0.565. It
persisted in the Title-and-length ($\pm10\%$) subset, at 0.856 versus
0.678, and under the other length bands.

June 2025 contained only 879 articles, all from 8 June. Tone MAE was
1.99 and implemented GCAM presence F1 was 0.820. The scheduled 23 June
archives were unavailable, which coincides with a gap in GDELT's
published files from 14 June to 1 July (the cause is not documented by
GDELT). The June estimates therefore describe a partial month. The
months after the outage (July--December 2025) were similar to the
preceding period. We therefore did not treat them as a separate period.

\begin{figure*}[!b]
  \centering
  \includegraphics[width=\textwidth]{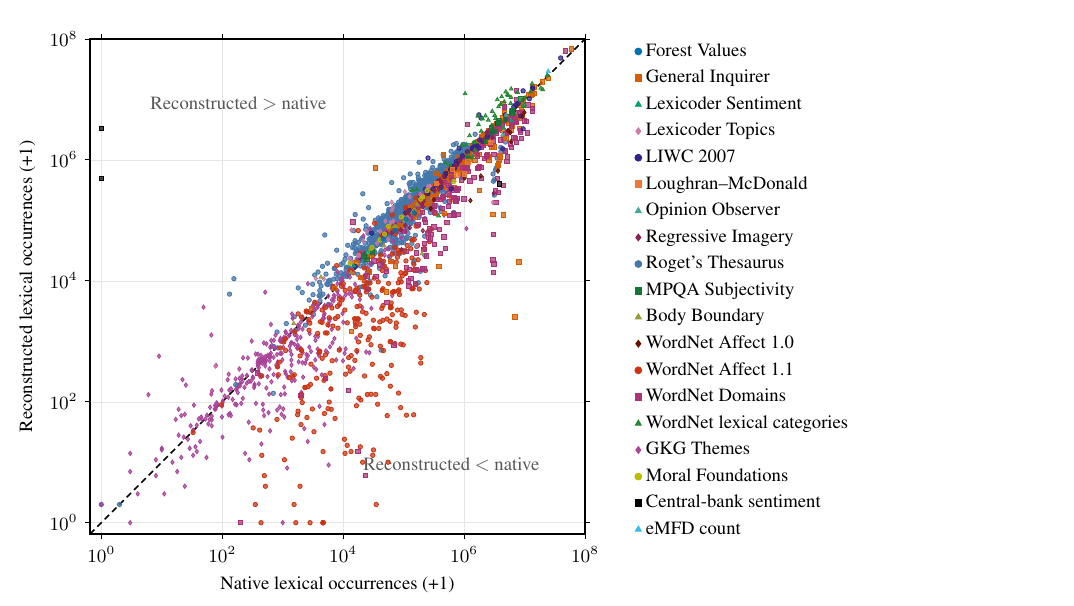}
  \caption{Pooled GCAM lexical counts by category. The dashed line is
    identity; logarithmic axes add one to retain zeros.}
  \label{fig:gcam-counts}
\end{figure*}

\subsection{Geographic Resources}
\label{sec:results-geography}
The larger gazetteer substantially increased inventory coverage but made
little difference to location identity F1
(Table~\ref{tab:geographic-agreement}). Expanding from 17,177 to 119,733
entries increased coverage of native within-article identities from
89.5\% to 97.1\%. The paired full-corpus location-F1 difference was
0.0006 [$-0.0002$, 0.0014]; both point estimates round to 0.654.
Complete count-tuple F1 increased from 0.066 to 0.070, a paired
difference of 0.0039 [0.0025, 0.0054]. The location contrast interval
includes zero, while the count-tuple interval is positive.

The Title-and-length ($\pm10\%$) subset showed the same direction of
change: location F1 was 0.725 with the bundled resource and 0.727 with
the expanded resource, while count-tuple F1 was 0.068 and 0.073. The
coverage column in Table~\ref{tab:geographic-agreement} describes the
full corpus for both subsets.

\begin{table}[H]
  \caption{Geographic Agreement}
  \label{tab:geographic-agreement}
  \centering\footnotesize
  \setlength{\tabcolsep}{2pt}
  \begin{tabular*}{\columnwidth}{@{\extracolsep{\fill}}lrrr@{}}
    \toprule
    & Coverage (\%) & Location F1 & Count full-tuple F1 \\
    \midrule
    BF & 89.5 & 0.654 [0.646, 0.661] & 0.066 [0.062, 0.072] \\
    BT & 89.5 & 0.725 [0.719, 0.729] & 0.068 [0.063, 0.074] \\
    EF & 97.1 & 0.654 [0.646, 0.662] & 0.070 [0.066, 0.076] \\
    ET & 97.1 & 0.727 [0.722, 0.731] & 0.073 [0.067, 0.078] \\
    \bottomrule
  \end{tabular*}
  \ArticleNote{BF: Bundled, Full corpus; BT: Bundled, Title-and-length
    ($\pm10\%$) subset; EF: Expanded, Full corpus; ET: Expanded,
    Title-and-length ($\pm10\%$) subset.
    Coverage: native identities covered in the full corpus. Micro F1 with
    95\% publisher-cluster intervals in brackets.}
\end{table}

\subsection{GCAM}
\label{sec:results-gcam}
GCAM lexical-match category presence had micro F1 0.824 [0.819, 0.827]
across the full observed inventory and 0.840 [0.836, 0.844] within the
implemented inventory (Table~\ref{tab:english-field-agreement}).
Category-macro F1 was lower, at 0.651 across 2,442 observed categories
and 0.674 across 2,359 implemented observed categories. Implemented
article-macro F1 was 0.826 [0.821, 0.830]. The profile implements 2,492
categories across all evaluated languages, including 2,372 English
categories. The English count comparison observed 2,442 categories:
2,359 implemented categories and 83 outside that inventory. Of the
implemented observed categories, 2,330 had positive native count support
and are plotted; the other 29 belong to families without native count
support. Categories absent from both count outputs are omitted from
these observed totals.

Dictionary-level results varied substantially
(Table~\ref{tab:gcam-agreement}, Appendix~\ref{app:gcam-results}).
General Inquirer~\cite{stone1966inquirer},
LIWC~\cite{pennebaker2007liwc}, and MPQA~\cite{wilson2005polarity} had
presence micro F1 of 0.927, 0.948, and 0.972, but count WAPE of 41.4\%,
38.2\%, and 36.3\%, respectively. Roget~\cite{roget1911thesaurus} had
micro F1 0.818 and WAPE 70.8\%; WordNet Affect
1.1~\cite{strapparava2004affect} had micro F1 0.320 and category-macro
F1 0.253. Thus recovering a category's presence did not imply agreement
on the number of lexical matches.

Fig.~\ref{fig:gcam-counts} shows distinct patterns across dictionary
families. The plot covers 2,330 categories in 19 families with positive
native count support. This view describes total lexical volume.
Reconstructed counts were generally lower for WordNet Affect, WordNet
Domains, and GKG Themes, but higher for Roget, LIWC, and WordNet lexical
categories. General Inquirer and MPQA showed differences in both
directions. Article-level overcounts and undercounts can cancel before
pooling.

The full and implemented presence scores include 153,866
reconstructed-only article--category assignments from weighted families
without positive native count support. Appendix~\ref{app:gcam-results}
explains their interpretation. For the VADER
lexicon~\cite{hutto2014vader}, paired mean lexical valence had MAE 0.24
[0.23, 0.25] and CCC 0.869 [0.855, 0.882] across 191,124 articles;
score-availability F1 was 0.994.

\subsection{Multilingual Results}
\label{sec:results-multilingual}
Tone of translated articles had higher MAE and lower CCC in every
language than in the English corpus (Table~\ref{tab:multilingual-tone}).
Translated-tone MAE ranged from 2.10 for Spanish to 2.64 for Chinese;
CCC ranged from 0.446 for German to 0.630 for Spanish. These values
compare with MAE 1.79 and CCC 0.749 for English, although the language
samples have different inclusion rules and article compositions.

\begin{table}[H]
  \caption{Translated Tone}
  \label{tab:multilingual-tone}
  \centering\footnotesize
  \setlength{\tabcolsep}{2pt}
  \begin{tabular*}{\columnwidth}{@{\extracolsep{\fill}}lrrr@{}}
    \toprule
    & Paired $n$ & MAE & CCC \\
    \midrule
    EN* & 193,550 & 1.79 [1.76, 1.82] & 0.749 [0.736, 0.759] \\
    FR & 500 & 2.40 [2.22, 2.58] & 0.515 [0.443, 0.581] \\
    ES & 500 & 2.10 [1.95, 2.27] & 0.630 [0.569, 0.681] \\
    DE & 500 & 2.61 [2.39, 2.84] & 0.446 [0.332, 0.543] \\
    RU & 500 & 2.56 [2.34, 2.80] & 0.593 [0.524, 0.657] \\
    ZH & 499 & 2.64 [2.37, 2.94] & 0.513 [0.402, 0.620] \\
    \bottomrule
  \end{tabular*}
  \ArticleNote{EN*: Full corpus, English comparison.
    Other rows: English translations. Incomplete Chinese translation excluded.
    Brackets: 95\% publisher-cluster intervals.}
\end{table}

Original-language GCAM is computed on the original text, not on
translations, and is therefore reported separately
(Table~\ref{tab:multilingual-agreement}). Original-language GCAM
presence micro F1 was 0.613 for French, 0.771 for Spanish, and 0.679 for
German; category-macro F1 was 0.509, 0.756, and 0.527, respectively.
Russian and Chinese each had one implemented Hedonometer category,
native presence was 377/500 for Russian and 500/500 for Chinese; F1 is
not informative with a single category.

Hedonometer happiness CCC ranged from 0.464 for German to 0.902 for
Spanish, with 340--500 valid paired articles per language.

\begin{table}[H]
  \caption{Original-Language GCAM}
  \label{tab:multilingual-agreement}
  \centering\footnotesize
  \setlength{\tabcolsep}{2pt}
  \begin{tabular*}{\columnwidth}{@{\extracolsep{\fill}}lrrrr@{}}
    \toprule
    & \multicolumn{2}{c}{Presence F1} & \multicolumn{2}{c}{Happiness} \\
    \cmidrule(lr){2-3}\cmidrule(l){4-5}
    & Micro & Macro & MAE & CCC \\
    \midrule
    FR & 0.613 [0.585, 0.639] & 0.509 & 0.039 [0.033, 0.044] & 0.826 [0.765, 0.878] \\
    ES & 0.771 [0.736, 0.803] & 0.756 & 0.040 [0.035, 0.045] & 0.902 [0.869, 0.926] \\
    DE & 0.679 [0.646, 0.712] & 0.527 & 0.083 [0.076, 0.093] & 0.464 [0.389, 0.542] \\
    RU\textsuperscript{a} & 0.860 [0.833, 0.886] & 0.860 & 0.060 [0.054, 0.067] & 0.802 [0.753, 0.843] \\
    ZH\textsuperscript{b} & 1.000 & 1.000 & 0.098 [0.087, 0.112] & 0.671 [0.578, 0.742] \\
    \bottomrule
  \end{tabular*}
  \ArticleNote{Implemented inventory; macro F1 averaged over categories.
    Paired happiness $n$: FR 372, ES 340, DE 402, RU 377, ZH 500.
    Brackets: 95\% publisher-cluster intervals. \textsuperscript{a,b}One category; see Section~\ref{sec:results-multilingual}.}
\end{table}

\section{Discussion}
\label{sec:discussion}
The main message of the results is that a reconstructed GKG record
cannot be judged as a whole. Agreement with native records is high
enough for some fields to be useful, partial for others, and too low for
the rest. We discuss what this means for users, why agreement differs
between fields, how it depends on input correspondence and time, and
which limitations bound these conclusions.

\subsection{What the Agreement Means for Users}
\label{sec:discussion-implications}
Reconstructed tone follows native tone moderately closely on average,
but individual articles can deviate considerably. Positive and negative
errors partly cancel, so a small mean bias does not imply close
agreement for single articles. Tone is therefore most suitable for
analyses that aggregate many articles, and less suitable for comparing
individual articles or detecting small differences near the neutral
band. Even for aggregates, a systematic bias does not average out, so
users should check it within their own groups and time periods.

Person and location identities are recovered moderately well, and
organizations less well. For persons, reconstructed outputs contained
more items than the native records, whereas for organizations they more
often omitted native items. Dates showed the same pattern as persons.
These are differences from native outputs, not measured factual errors,
so extra reconstructed items and omitted native items cannot be called
mistakes from this comparison alone.

Agreement dropped whenever a field required more than a matching value.
A number is easy to match, but matching what it describes and where it
occurred is much harder. Analyses that depend on correct objects and
location attachments therefore cannot rely on reconstructed counts and
amounts. Likewise, low agreement on quotation texts means that
reconstructed records are not a substitute when an analysis needs the
same quoted passages as native GKG.

\subsection{Why Agreement Differs Between Fields}
\label{sec:discussion-field-differences}
Coverage and agreement answer different questions. A larger gazetteer
made far more native places available, yet location agreement barely
changed. A place can be in the resource without being proposed by entity
recognition, selected by the resolver, or attached to the same count.
Our comparison changed the inventory, the aliases, and the historical
frequencies together, so it shows that coverage alone does not drive
agreement but cannot separate these factors. The same distinction
applies to themes: the result for implemented labels describes only the
part of the inventory we reconstruct and says nothing about the native
labels we do not.

For GCAM, recovering that a category occurs is easier than reproducing
how often its words occur, and reproducing a weighted score is a
separate question again. Presence agreement was high for many
dictionaries even where lexical counts differed substantially. Weighted
dictionaries must be judged by their paired scores rather than by
presence, because native records often contain no corresponding count.
The summary measures also differ in what they reward: common categories
dominate micro F1, while category-macro F1 gives equal weight to rare
ones. Neither replaces the dictionary-level results.

Tone computed on English translations was less accurate than for English
articles. This difference combines language, sample composition,
retrieval, and differences between translations and those processed by
GDELT, and we cannot attribute it to any one of them. Original-language
GCAM relies on different dictionaries for each language, so it should
not be compared across languages.

\subsection{Input Correspondence}
\label{sec:discussion-input-correspondence}
Agreement was higher in subsets where retrieved and native titles and
lengths matched, which is consistent with part of the disagreement
coming from differences in the text. Most disagreement remained,
however, and the results were broadly across length bands and
near-duplicate thresholds. These subsets are selected populations, not
controlled experiments, and several of them favor agreement for reasons
unrelated to text identity. Tone and the length filter share token
counts, so selecting similar lengths mechanically favors tone. Entity
matching does not share this denominator, but selection may still raise
entity agreement. The quotation-anchor subset contains articles with
several long quotations, which is a distinct type of article. Because
titles and lengths do not establish that the bodies are identical, we
cannot separate the effect of input differences from extraction
differences, and we cannot assign entity disagreement to recognition,
normalization, or entity-type definitions.

\subsection{Temporal Variation}
\label{sec:discussion-temporal}
Agreement showed no clear improvement toward more recent records, so
there is no sign that the tone word lists, which were tuned on 2026
data, fit recent articles better. GCAM count disagreement rose in
2021--2022 across several dictionaries, but only in the full corpus and
the Title-matched subset, not in the length-matched subsets. Because
lexical counts scale with text length, this suggests the rise reflects
differences in the retrieved text rather than a change in GDELT's
processing. Length matching also reduces count error by construction, so
we cannot confirm this. Date agreement varied by calendar month in every
subset, so the length restriction does not explain it. Because each
month is represented by only 12 sampled days, the pattern may also
reflect those days. We did not investigate it further.

\subsection{Limitations}
\label{sec:discussion-limitations}
Retrospective retrieval embeds survivorship into the article corpus.
Archive availability, access restrictions, extraction checks, language
verification, and deduplication determine which pairs enter the
evaluation, so the results describe the retained pairs rather than all
GKG-linked news. The annual retention fractions combine several kinds of
loss and do not isolate link rot.

The gazetteer and the tone word lists were derived from native GKG
records outside the evaluation period, so the reconstruction is not
independent of GDELT's output. Tone agreement is evaluated on data
separate from the development corpus, but it reflects calibration as
well as method.

Some resources differ from the native procedures, which are unknown:
dictionary editions, our authored theme rules, text preprocessing, and
the approximate French imagery list. Fields other than tone and GCAM
were evaluated for English only, and each non-English language is
represented by a small sample. Publisher-cluster intervals describe
resampling variation within the retained corpus; they do not correct for
any of the above or for measurement uncertainty.

We measured agreement but did not diagnose its causes. Explaining the
weaker fields, such as organizations, dates, and counts, would require
manual error analysis, and a clean separation of text differences from
extraction differences would require the original article bodies.

\section{Conclusion}
\label{sec:conclusion}
The study provides a baseline for reconstructing GKG fields using
lexical resources, extraction rules, and native-derived resources. It
evaluates 193,550 English articles and 2,500 articles in five additional
languages. Agreement varies across fields, matching units, and
implemented inventories. The Title-and-length ($\pm10\%$) subset has
lower tone error and higher entity F1. A larger gazetteer increased
geographic coverage, but this did not produce a comparable gain in
location agreement. Original-language GCAM and translated tone require
separate interpretation. Person and location identities provide partial
recovery of native content. Agreement did not improve with more recent
records, but it varied over time in ways that depend on input
correspondence. Complete counts, quotation texts, and dates require
further development and validation before they can be used in place of
the native fields. For tone, the observed error scale should be compared
with the differences an application seeks to measure.

Future work should diagnose the weaker fields through manual error
analysis, quantify the sensitivity of each field to text differences by
perturbing and re-extracting article bodies, and collect article text
promptly after GKG publication to obtain closer input correspondence.

\section{Data and Code Availability}
\label{sec:data-availability}
The \software\ software, including the benchmark used for the
evaluation, is available at PyPI\footnote{\pypi} or
GitHub\footnote{\github}. The native GKG records are available from
GDELT's public archives. The retrieved article texts cannot be
redistributed.

\appendices
\setcounter{table}{0}
\renewcommand{\thetable}{\Alph{section}\arabic{table}}
\renewcommand{\theHtable}{appendix.\Alph{section}.\arabic{table}}

\section{GCAM Resources and Matching}
\label{app:resources}

\subsection{Lexical-match Dictionaries}
\label{app:resources-lexical-dictionaries}
English GCAM resources included Forest Values~\cite{bengston1995forest},
General Inquirer~\cite{stone1966inquirer}, Lexicoder
Sentiment~\cite{young2012affective}, Lexicoder
Topics~\cite{albaugh2013topics},
Loughran--McDonald~\cite{loughran2011liability}, Opinion
Observer~\cite{hu2004mining}, Regressive
Imagery~\cite{martindale1987narrative}, Body
Boundary~\cite{wilson2006boundary}, Moral
Foundations~\cite{grahamhaidtMFD}, and central-bank
sentiment~\cite{correa2017stability} dictionaries. General Inquirer's
228 categories combined its spreadsheet vocabulary with 11 categories
imported from its dictionary text; LIWC supplied 62 GCAM-compatible
categories. MPQA supplied 14 categories defined by subjectivity
strength, prior polarity, and their combinations. Lexicoder Sentiment
used the August 2015 category files, Lexicoder Topics used the June 2013
dictionary, Loughran--McDonald used active entries in the 1993--2025
master dictionary, and central-bank sentiment used the 2017
financial-stability-report dictionary.

\subsection{Semantic Dictionaries}
\label{app:resources-semantic-dictionaries}
Resources also included Roget's 1911
thesaurus~\cite{roget1911thesaurus}, WordNet Affect 1.0 and
1.1~\cite{strapparava2004affect}, WordNet Domains
3.2~\cite{magnini2000domains}, and WordNet 3.1 lexical
categories~\cite{miller1995wordnet}. The Affect mappings used WordNet
1.6 synsets, whereas Domains used WordNet 2.0. Vocabulary was formed
from the union of annotated senses, without contextual word-sense
disambiguation. Affect used direct category annotations, and Domains did
not propagate unannotated parent categories. Roget and the WordNet
lexical-match dictionaries used WordNet morphological alternatives
without part-of-speech disambiguation.

\subsection{Weighted Dictionaries}
\label{app:resources-weighted-dictionaries}
The English scoring resources were SentiWordNet
3.0~\cite{baccianella2010sentiwordnet},
SentiWords~\cite{guerini2013prior}, ANEW~\cite{bradley1999anew},
ML-SENTICON~\cite{cruz2014building},
Hedonometer~\cite{dodds2011hedonometer}, VADER~\cite{hutto2014vader},
and the Extended Moral Foundations Dictionary
(eMFD)~\cite{hopp2020emfd}. English ANEW contributed 9 rating columns.
ML-SENTICON contributed 8 cumulative layers separately for positive and
negative polarity. Hedonometer used the English labMT word list. VADER
used the 3.3.2 lexicon. The GCAM VADER score was a mean lexical valence,
without sentence-level negation, emphasis, or compound-score rules. eMFD
contributed one moral-word count and ten score-only categories.

\subsection{Language-specific Coverage}
\label{app:resources-language-coverage}
French (FR) bodies used the French Regressive Imagery
dictionary~\cite{hogenraadRIDFrench}, and German (DE) bodies used its
German counterpart~\cite{delphendahlRIDGerman}. Spanish (ES) bodies used
Spanish ANEW~\cite{redondo2007anew} and Spanish
ML-SENTICON~\cite{cruz2014building}. Spanish ANEW contributed 12 rating
columns, including familiarity, concreteness, and imagery; unavailable
ratings remained missing. The multilingual Hedonometer
lists~\cite{dodds2014a} supplied one happiness category in each of
French, Spanish, German, Russian (RU), and Chinese (ZH). French
additionally used the imagery dictionary~\cite{hogenraad1981imagery}
with the approximation described below. Original-language Russian and
Chinese coverage was limited to Hedonometer. English (EN) translations
were evaluated with the English inventory, separately from these
original-language measures.

\subsection{French Imagery Approximation}
\label{app:resources-french-imagery}
The French imagery resource~\cite{hogenraad1981imagery} was constructed
from nouns assigned to six published rating intervals. Line-by-line
Tesseract OCR~\cite{smith2007tesseract} with French recognition data
recovered candidate nouns and interval membership. Each retained noun
received its interval midpoint on the GCAM 10--70 scale: 19.5, 29.5,
39.5, 49.5, 59.5, or 67.5. The conversion extracted 1,127 entries and
retained 1,120 distinct scored words after excluding unusable multiword
entries and conflicting interval assignments. Midpoint substitution
approximates the scores because individual noun means cannot be
recovered. Matched nouns were combined using the same
occurrence-weighted mean as the other weighted resources. Its count and
score outputs represented one approximate category.

\begin{table*}[!b]
\setcounter{table}{0}
\renewcommand{\thetable}{\getrefnumber{app:gcam-results}\arabic{table}}
\renewcommand{\theHtable}{appendix.\getrefnumber{app:gcam-results}.\arabic{table}}
\caption{English GCAM Agreement}
\label{tab:gcam-agreement}
\centering\footnotesize
\setlength{\tabcolsep}{4pt}
\begin{tabular*}{\textwidth}{@{\extracolsep{\fill}}lrrrrr@{}}
\toprule
Dictionary / inventory & Observed categories & Native assignments & Micro F1 & Category-macro F1 & WAPE (\%) \\
\midrule
GCAM (all observed) & 2,442 & 178,070,983 & 0.824 & 0.651 & --- \\
GCAM (implemented) & 2,359 & 171,168,198 & 0.840 & 0.674 & --- \\
Forest Values~\cite{bengston1995forest} & 4 & 148,939 & 0.857 & 0.858 & 38.6 \\
General Inquirer~\cite{stone1966inquirer} & 228 & 31,872,896 & 0.927 & 0.893 & 41.4 \\
Lexicoder Sentiment~\cite{young2012affective} & 4 & 412,145 & 0.971 & 0.922 & 42.7 \\
Lexicoder Topics~\cite{albaugh2013topics} & 28 & 857,526 & 0.638 & 0.626 & 95.8 \\
LIWC 2007~\cite{pennebaker2007liwc} & 62 & 9,018,002 & 0.948 & 0.928 & 38.2 \\
Loughran--McDonald~\cite{loughran2011liability} & 6 & 905,545 & 0.885 & 0.879 & 44.8 \\
Opinion Observer~\cite{hu2004mining} & 2 & 378,751 & 0.977 & 0.977 & 45.1 \\
Regressive Imagery~\cite{martindale1987narrative} & 43 & 3,850,148 & 0.876 & 0.843 & 44.9 \\
Roget (1911)~\cite{roget1911thesaurus} & 1,042 & 79,170,672 & 0.818 & 0.760 & 70.8 \\
MPQA Subjectivity~\cite{wilson2005polarity} & 14 & 2,087,106 & 0.972 & 0.944 & 36.3 \\
Body Boundary~\cite{wilson2006boundary} & 14 & 539,759 & 0.754 & 0.742 & 56.2 \\
WordNet Affect 1.0~\cite{strapparava2004affect} & 11 & 1,877,150 & 0.888 & 0.857 & 61.4 \\
WordNet Affect 1.1~\cite{strapparava2004affect} & 279 & 6,667,993 & 0.320 & 0.253 & 93.1 \\
WordNet Domains~\cite{magnini2000domains} & 168 & 23,030,289 & 0.815 & 0.667 & 62.9 \\
WordNet lexical categories~\cite{miller1995wordnet} & 44 & 7,690,347 & 0.962 & 0.951 & 76.0 \\
GKG Themes~\cite{leetaru2013gkg} & 366 & 1,681,477 & 0.664 & 0.543 & 68.5 \\
Moral Foundations~\cite{grahamhaidtMFD} & 11 & 613,125 & 0.882 & 0.865 & 42.7 \\
Central-bank sentiment~\cite{correa2017stability} & 3 & 187,741 & 0.384 & 0.259 & 189.6 \\
eMFD count~\cite{hopp2020emfd} & 1 & 178,587 & 0.960 & 0.960 & 34.1 \\
ANEW~\cite{bradley1999anew} & 9 & 0 & 0.000 & 0.000 & --- \\
ML-SENTICON~\cite{cruz2014building} & 16 & 0 & 0.000 & 0.000 & --- \\
SentiWordNet~\cite{baccianella2010sentiwordnet} & 2 & 0 & 0.000 & 0.000 & --- \\
SentiWords~\cite{guerini2013prior} & 1 & 0 & 0.000 & 0.000 & --- \\
VADER~\cite{hutto2014vader} & 1 & 0 & 0.000 & 0.000 & --- \\
\bottomrule
\end{tabular*}
\ArticleNote{WAPE measures lexical counts. A dash denotes an undefined or unreported value.}
\end{table*}
\setcounter{table}{0}

\subsection{Theme Approximation}
\label{app:resources-themes}
GKG Themes~\cite{leetaru2013gkg} contributed 367 categories: 99
source-backed categories and 268 categories implemented through authored
rules. The source-backed rules included vocabulary from
CrisisLex~\cite{olteanu2014crisislex}. Authored rules associated literal
expressions and aliases with public GKG labels; for example,
\emph{austerity} and \emph{spending cuts} mapped to \texttt{AUSTERITY},
while \emph{property rights} and \emph{land ownership rights} mapped to
\texttt{PROPERTY\_RIGHTS}. Existing rules and document-level context
requirements and exclusions were retained. Theme counts used distinct
category--start--end matches, including overlapping spans. The authored
rules provided an operational vocabulary for the named categories. All
268 categories were marked as approximations. The remaining native theme
category, \texttt{TAX\_SPECIALDEATH}, was not implemented.

Table~\ref{tab:gcam-resources} gives the cited inventory of lexical
resources and GKG Themes rules across the evaluated languages.
Appendix~\ref{app:gcam-results} reports the English dictionary-level
presence and count results.

\begin{table}[!t]
  \caption{GCAM Dictionaries and Implemented Coverage}
  \label{tab:gcam-resources}
  \centering\footnotesize
  \setlength{\tabcolsep}{2pt}
  \begin{tabularx}{\columnwidth}{@{}rXrrr@{}}
    \toprule
    ID & Resource & Cat. & Counts & Scores \\
    \midrule
    1 & Forest Values~\cite{bengston1995forest} & 4 & 4 & 0 \\
    2 & General Inquirer~\cite{stone1966inquirer} & 228 & 228 & 0 \\
    3 & Lexicoder Sentiment~\cite{young2012affective} & 4 & 4 & 0 \\
    4 & Lexicoder Topics~\cite{albaugh2013topics} & 28 & 28 & 0 \\
    5 & LIWC 2007~\cite{pennebaker2007liwc} & 62 & 62 & 0 \\
    6 & Loughran--McDonald~\cite{loughran2011liability} & 6 & 6 & 0 \\
    7 & Opinion Observer (Hu--Liu)~\cite{hu2004mining} & 2 & 2 & 0 \\
    8 & Regressive Imagery~\cite{martindale1987narrative} & 43 & 43 & 0 \\
    9 & Roget's Thesaurus (1911)~\cite{roget1911thesaurus} & 1,042 & 1,042 & 0 \\
    10 & SentiWordNet 3.0~\cite{baccianella2010sentiwordnet} & 2 & 2 & 2 \\
    11 & SentiWords~\cite{guerini2013prior} & 1 & 1 & 1 \\
    12 & MPQA Subjectivity~\cite{wilson2005polarity} & 14 & 14 & 0 \\
    13 & Body Boundary~\cite{wilson2006boundary} & 14 & 14 & 0 \\
    14 & WordNet Affect 1.0~\cite{strapparava2004affect} & 11 & 11 & 0 \\
    15 & WordNet Affect 1.1~\cite{strapparava2004affect} & 280 & 280 & 0 \\
    16 & WordNet Domains 3.2~\cite{magnini2000domains} & 168 & 168 & 0 \\
    17 & WordNet 3.1 lexical categories~\cite{miller1995wordnet} & 44 & 44 & 0 \\
    18 & GKG Themes\textsuperscript{a}~\cite{leetaru2013gkg} & 367 & 367 & 0 \\
    19 & ANEW~\cite{bradley1999anew} & 9 & 9 & 9 \\
    20 & ML-SENTICON~\cite{cruz2014building} & 16 & 16 & 16 \\
    21 & Hedonometer~\cite{dodds2011hedonometer} & 1 & 1 & 1 \\
    22 & ANEW (ES)~\cite{redondo2007anew} & 12 & 12 & 12 \\
    23 & ML-SENTICON (ES)~\cite{cruz2014building} & 16 & 16 & 16 \\
    24 & Hedonometer (multilingual)~\cite{dodds2014a} & 5 & 5 & 5 \\
    25 & Moral Foundations~\cite{grahamhaidtMFD} & 11 & 11 & 0 \\
    26 & VADER lexicon~\cite{hutto2014vader} & 1 & 1 & 1 \\
    28 & Regressive Imagery (FR)~\cite{hogenraadRIDFrench} & 43 & 43 & 0 \\
    33 & Regressive Imagery (DE)~\cite{delphendahlRIDGerman} & 43 & 43 & 0 \\
    38 & French imagery (FR)\textsuperscript{b}~\cite{hogenraad1981imagery} & 1 & 1 & 1 \\
    41 & Central-bank sentiment~\cite{correa2017stability} & 3 & 3 & 0 \\
    42 & eMFD~\cite{hopp2020emfd} & 11 & 1 & 10 \\
    \midrule
    \multicolumn{2}{@{}l}{\textbf{Total}} & \textbf{2,492} & \textbf{2,482} & \textbf{74} \\
    \bottomrule
  \end{tabularx}
  \ArticleNote{EN unless marked; multilingual: FR/ES/DE/RU/ZH.
    \textsuperscript{a,b}Approximations.}
\end{table}

\subsection{Matching and Aggregation}
\label{app:resources-matching}
Category names were mapped to GCAM identifiers using explicit crosswalks
and spelling aliases. English matching used the tone tokenizer;
non-English matching preserved Unicode letters, combining marks, and
internal apostrophes and hyphens. Chinese used Jieba~\cite{sun2020jieba}
segmentation with its standard dictionary and new-word inference
disabled. Literal and wildcard lexical-match dictionaries selected the
longest non-overlapping match within each category, with categories
evaluated independently. Source exclusions masked matching tokens while
preserving boundaries. French and German Regressive Imagery used their
supplied wildcards without additional inferred stemming. VADER used
whitespace matching to preserve recognized emoticons. For the
rating-table imports, duplicate source records were removed, distinct
records for the same normalized token were averaged, and unhandled
multiword or punctuation-split entries were omitted.

For article $i$ and weighted category $k$, let $f_i(w)$ denote the
frequency of a matched token and $s_k(w)$ its imported weight. The
reconstructed score was

\begin{equation}
    \label{eq:reconstructed-gcam}
    \widehat{y}_{ik}=\frac{\sum_{w\in M_{ik}}f_i(w)s_k(w)}{\sum_{w\in M_{ik}}f_i(w)},
\end{equation}
where $M_{ik}$ contains matched tokens with an available weight for that
category. Thus scores were averaged over matched occurrences. A category
with no scored matches had no reconstructed value; a matched mean of
zero remained a valid value. Count outputs represented lexical match
frequencies.

\subsection{Coverage Accounting}
\label{app:resources-coverage}
The evaluation used 2,492 distinct categories: 2,372 English, 45 French,
29 Spanish, 44 German, one Russian, and one Chinese. Of these, 2,418
emitted counts only, 64 emitted both counts and scores, and 10 eMFD
categories emitted scores only. The inventory therefore supplies 2,482
count outputs and 74 weighted outputs, with 64 categories shared between
them. It includes 2,223 source-backed categories and 269 approximations.
The 181 resource categories assigned to other languages were outside
this evaluation. Unimplemented native categories remain in full native
presence comparisons but are excluded from implemented comparisons.

\setcounter{table}{1}

\section{GCAM Dictionary-Level Results}
\label{app:gcam-results}
Table~\ref{tab:gcam-agreement} reports the full-English dictionary
results within the implemented inventory. Presence requires a positive
lexical count; the existence of a weighted score does not affect it. If
a native weighted value has no companion count, that article--category
count comparison is omitted on both sides. In articles without such a
native weighted value, a remaining reconstructed-only count can still
produce a false positive. A family can consequently have zero
lexical-match presence F1 despite informative paired weighted-score
agreement. Rows with zero native count support do not measure
weighted-score accuracy. Weighted-score MAE and CCC used only available
pairs.

\clearpage
\ArticleReferences
\end{document}